\documentclass[runningheads]{llncs}

\usepackage{eccv}

\usepackage{eccvabbrv}

\usepackage{graphicx}
\usepackage{booktabs}

\usepackage[accsupp]{axessibility}  

\usepackage{color}
\usepackage{wrapfig}
\usepackage{array}

\usepackage{textcomp}
\usepackage{multirow}
\usepackage{hhline}

\usepackage{algorithmic}
\usepackage{microtype}
\usepackage{mathtools}
\usepackage{booktabs} %
\usepackage[ruled]{algorithm2e} %
\usepackage{comment}
\usepackage{graphicx}

\usepackage{enumitem}
\usepackage{balance}
\usepackage[normalem]{ulem} 
\usepackage{amssymb}
\usepackage[tone, extra, safe]{tipa}
\usepackage{tipx}
\usepackage{amssymb}
\usepackage{enumitem}
\usepackage[dvipsnames]{xcolor}
\usepackage{arydshln}

\usepackage{adjustbox}
\usepackage{makecell}

\usepackage{pifont}

\usepackage{listings}

\definecolor{codegreen}{rgb}{0,0.6,0}
\definecolor{codegray}{rgb}{0.5,0.5,0.5}
\definecolor{codepurple}{rgb}{0.58,0,0.82}
\definecolor{backcolour}{rgb}{0.95,0.95,0.92}

\lstdefinestyle{mystyle}{
    backgroundcolor=\color{backcolour},   
    commentstyle=\color{codegreen},
    keywordstyle=\color{magenta},
    numberstyle=\tiny\color{codegray},
    stringstyle=\color{codepurple},
    basicstyle=\ttfamily\footnotesize,
    breakatwhitespace=false,         
    breaklines=true,                 
    captionpos=b,                    
    keepspaces=true,                 
    numbers=left,                    
    numbersep=5pt,                  
    showspaces=false,                
    showstringspaces=false,
    showtabs=false,                  
    tabsize=2
}

\usepackage[most]{tcolorbox} 

\tcbset{
  colback=gray!20, 
  colframe=black,  
  width=\textwidth, 
  boxrule=0pt, 
  arc=0mm, 
  left=2mm, 
  right=2mm,
  top=1mm,
  bottom=1mm,
  boxsep=0mm,
}

\usepackage{fancyvrb}

\newcommand{\aref}[1]{Appendix~\ref*{#1}}

\newcommand{\modelname}{\textcolor{black}{MOCO}\xspace}
\newcommand{\modelnames}{\textcolor{black}{MOCO's}\xspace}

\definecolor{DeltaColor}{rgb}{0.039,0.73,0.71}
\definecolor{SigmaColor}{rgb}{0.98,0.45,0.0}
\definecolor{AlphaColor}{rgb}{0,0,0.8}
\definecolor{BetaColor}{rgb}{0.8,0,0.8}
\definecolor{GammaColor}{rgb}{0.514,0.34,0.224}
\definecolor{EpsilonColor}{rgb}{0.353,0.725,0.906}
\definecolor{PurpleColor}{rgb}{0.5,0,0.7}
\definecolor{OrangeColor}{rgb}{0.914,0.541,0.141}
\definecolor{GreenColor}{rgb}{0.137,0.573,0.565}
\definecolor{RedColor}{rgb}{0.949,0.275, 0.224}
\definecolor{LightCyan}{rgb}{0.88,1,1}
\definecolor{Gray}{gray}{0.3}
\definecolor{Strawberry}{rgb}{1,0.26,0.64}

\definecolor{BetaColor}{rgb}{0.8,0,0.8}

\newcommand{\rev}[1]{#1}

\definecolor{LightCyan}{rgb}{0.88,1,1}
\definecolor{lightgray}{rgb}{0.9,0.9,0.9}

\newcommand{\qheading}[1]{\noindent\textbf{#1}}

\definecolor{GreenColor}{rgb}{0.137,0.573,0.565}

\renewcommand{\paragraph}[1]{\medskip\noindent\textbf{#1}\ \ }

\makeatletter
\newcommand*\bigcdot{\mathpalette\bigcdot@{.5}}
\newcommand*\bigcdot@[2]{\mathbin{\vcenter{\hbox{\scalebox{#2}{$\m@th#1\bullet$}}}}}
\makeatother

\usepackage{hyperref}
\usepackage[capitalize]{cleveref}

\usepackage{orcidlink}

\begin{document}

\title{Multi-Modal Controlled Coherent Motion Generation} 


\author{Yifei Liu\inst{1,2} \and
Qiong Cao \inst{2} \and
Hongwei Yi \inst{3}  \and
Huaiguang Jiang \inst{1} \and \\
Changxing Ding \inst{1} \thanks{Corresponding author.}
}

\authorrunning{Y.~Liu et al.}

\institute{South China University of Technology \\
\email{ft\_lyf@mail.scut.edu.cn, hihuagong2021@scut.edu.cn, chxding@scut.edu.cn}
\and Joy Future Academy \\
\email{mathqiong2012@gmail.com} 
\and Peking University \\
\email{hongweiyi@pku.edu.cn} 
}

\maketitle

\begin{abstract}
It is natural for us to walk and talk simultaneously. This paper tackles the challenge of replicating such natural behaviors in 3D avatar motion generation driven by concurrent multi-modal inputs, \eg, a text description ``a man is walking'' alongside a speech audio.
Existing methods, constrained by the scarcity of aligned multi-modal data, typically combine motions from individual modalities sequentially or through weighted sum. However, they often result in mismatched or unrealistic movements. To overcome these limitations, we propose \textbf{\modelname}, a novel diffusion-based framework capable of processing multiple simultaneous inputs—including speech audio, text descriptions, and trajectory data—to generate coherent and lifelike motions without requiring \rev{aligned multi-modal data}.
Our key innovation lies in decoupling the motion generation process. During each denoising step, the diffusion model independently generates motions for each modality from the input noise and assembles the body parts according to predefined spatial rules. The resulting combined motion is then diffused and serves as the input noise for the subsequent denoising step. This iterative approach enables each modality to refine its contribution within the context of the overall motion, progressively harmonizing movements across modalities. Consequently, the generated motions become increasingly natural and fluid with each iteration, achieving coherent and synchronized behaviors.
We evaluate our approach using a purpose-built multi-modal benchmark. Experimental results demonstrate that \textbf{\modelname} outperforms existing baselines, advancing the field of multi-modal motion generation for 3D avatars.
The code will be released on \url{https://feifeifeiliu.github.io/MOCO}.
\keywords{Human Motion Generation \and Diffusion Models \and Multi-Modal}
\end{abstract}

\section{Introduction}
\label{sec:intro}

Imagine watching a virtual talk show where the host delivers engaging dialogue complemented by expressive gestures, natural body movements, and precise movement paths. The host walks across the stage following a scripted trajectory, uses hand gestures to emphasize points based on their speech, and shifts posture in response to both the conversation's flow and predefined text instructions—all occurring in perfect harmony. This level of realism transforms the viewing experience, making interactions feel genuine and immersive. Achieving such lifelike behavior is no small feat, yet it is essential for enhancing user engagement in applications ranging from virtual reality to interactive gaming and beyond.

Driving a 3D avatar to perform such lifelike motions involves managing multiple control signals, such as text descriptions, speech audio, and trajectory data. Multi-modal signals may be provided concurrently, for instance, a text prompt like ``a man is walking'' alongside a speech audio clip. However, most prior works primarily focus on single-modality control, such as text-to-motion \cite{guo2022humanml3d, tevet2022mdm} or speech-to-gesture \cite{ginosar2019learning, yi2023talkshow}. Recent studies \cite{zhou2023ude, Zhou2023AUF, zhang2024large} have explored designing unified models capable of addressing multi-modal signals by leveraging datasets from different generation tasks. Nevertheless, these models typically process only one modality at a time, combining motions conditioned on different inputs in a limited and sequential manner when multiple control signals are present.

\newcommand{\teaserCaption}{
\textbf{Examples of Multi-Modal Controlled Motion Generation.} Given multiple control signals of different modalities—including text descriptions, speech audio, and trajectory data—our \modelname framework generates realistic and coherent holistic body motion. This includes both body movements and detailed features such as facial expressions and hand gestures, all closely aligned with the provided conditions. To clearly illustrate this, we highlight two clips with temporal zoom, showcasing the natural integration of speech gestures and lower-body movements in our generated motions. Additional visual results are provided in \cref{fig:qualitative} and the \textbf{supplementary materials}.}

\begin{figure*}
    \centering
    
    \includegraphics[width=0.95\linewidth]{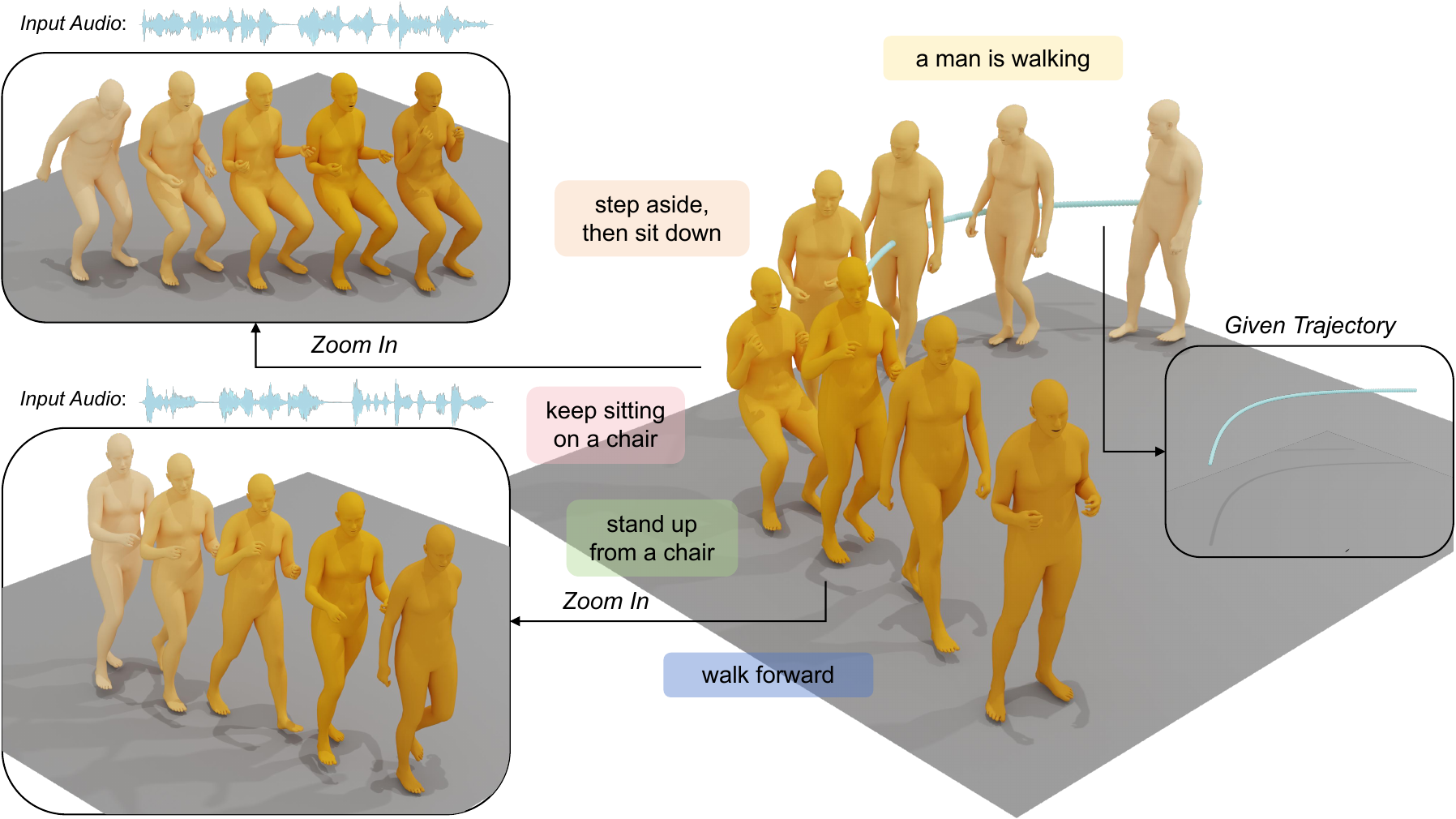}
    \caption{\teaserCaption}
    \label{fig:teaser}
\end{figure*}

The primary challenge in achieving simultaneous multi-modal control of motion generation is the scarcity of aligned multi-modal data. While collecting additional multi-modal data could help, it requires significant resources. In addition, the activity regions in speech-to-gesture datasets are often limited, making it hard to train models that can generate trajectory-controlled speech gestures. Some efforts address this issue by combining the predictions of text-driven model and audio-driven model through weighted sum \cite{yang2024freetalker, chen2024enabling}. \cite{ling2023mcm} suggested using speech scripts as pseudo text labels to create aligned text-audio-motion datasets, replacing scripts with movement descriptions during inference. However, these approaches face inherent limitations: for weighted sum, the predominance of standing poses in speech-to-gesture datasets creates imbalanced model corrections; for pseudo-labels, the use of speech scripts as training labels limits generalization to diverse motion descriptions during inference.
To overcome these challenges, we propose a novel diffusion-based framework, named \textbf{M}ulti-M\textbf{O}dal \textbf{C}ontrolled C\textbf{O}herent Motion Synthesis (\modelname).
MOCO exploits a natural spatial decomposition: speech audio primarily drives upper-body dynamics—--gestures and facial expressions--—while text descriptions mainly influence lower-body movements such as walking and stance shifts (see Appendix~A for supporting statistics).
To achieve this, MOCO is first trained on multiple datasets, ensuring that the model can independently generate motions conditioned on either text or speech inputs. 
At each denoising step, the model generates motions for each modality separately from the input noise and assembles the body parts according to predefined spatial rules, i.e. combining audio-driven upper-body motion with text-driven lower-body motion. This combined motion is then diffused and used as the input noise for the next denoising step. The separation ensures that each body part's motion is highly aligned with its corresponding input condition, while the iterative process conditions each generation step on the current state of the combined motion. This allows each modality to refine its contribution within the context of the overall movement. Consequently, with each iteration, the motions generated for different body parts become increasingly harmonized, resulting in natural movements that exhibit coherent and synchronized behaviors. Furthermore, this decoupled generation process enables our framework to incorporate trajectory control into co-speech motion generation. We can leverage trajectory data to generate text-driven motion and combine it with audio-driven motion, producing speech gestures that closely align with the given trajectory.

To facilitate the evaluation of this novel task, \textbf{we develop a multi-modal benchmark} comprising 1,000 test clips which are generated from 40 fundamental text descriptions of body movements (\eg, ``walk forwards'' and ``step back and sit down'') and 694 audio clips from eight different speakers. Each test clip integrates two text prompts describing a movement with two speech audio clips. We rigorously evaluate our approach against baseline methods using both text-to-motion and speech-to-gesture metrics. Experimental results demonstrate that our method outperforms existing baselines, advancing the field of multi-modal controlled motion generation for 3D avatars.
\section{Related Work}
\label{sec:rel_work}

\subsection{Multi-Modal Conditioned Motion Generation}
In recent years, human motion generation has received significant attention, largely driven by advancements in dataset collection. Various scenarios have been explored depending on the input conditions, including action labels \cite{guo2020action2motion}, text descriptions \cite{guo2022humanml3d, tevet2022mdm, zhang2022motiondiffuse, chen2023executing, harithas2024motionglot, wang2025fg, hosseyni2025bad, bae2025less, li2025lamp, zhang2025kinmo, hong2025salad}, speech audio \cite{ginosar2019learning, yi2023talkshow, liu2023emage, liu2024probtalk, zhang2024semtalk, liu2025gesturelsm, qi2025cocogesture, jiang2026smoothsync, sha20263dgespolicy}, music \cite{li2021aist, siyao2022bailando, tseng2023edge}, scene context \cite{hassan2019prox, ma2024richcat}, spatial signal \cite{xie2023omnicontrol, tevet2024closd}, and even the motion of another person \cite{Liu2023InteractiveHO, sun2025beyond,cen2025ready, wang2025timotion, gupta2026unified, peng2026dyadit}. Beyond single-modality control, research efforts have been extended to handle integrating multi-modal signals. For instance, some works considered speaker identity, speech audio, and transcripts to generate conversational gestures \cite{yoon2020trimodal, ao2022rhythmic, ao2023gesturediffuclip, xu2024mambatalk}, while other works jointly leverage text and scene inputs for motion generation \cite{cong2024laserhuman, cen2024generating, wang2024move, yi2024tesmo}. Moreover, some works focus on integrating various datasets to train unified motion models that enhance scalability and applicability across multiple scenarios \cite{zhou2023ude, zhang2024large, chen2024body_of_language, zhang2025motion, hu2025hmvlm}. 

A critical limitation persists, however: current models depend heavily on aligned multi-modal training data, limiting their ability to handle novel input combinations (\eg, text with audio, audio with trajectory) unseen in training. To address this, FreeTalker \cite{yang2024freetalker} and SynTalker~\cite{chen2024enabling}\rev{---the latter concurrent with the first version of this paper~\cite{liumulti}---}propose combining predictions from text-driven and audio-driven models through weighted sum. Liu et al. \cite{ling2023mcm} suggested using speech scripts as pseudo text labels (\eg, ``A male speaker is saying: `I am shocked by what you have done.' '') to create aligned text-audio-motion datasets for training and replace these scripts with movement descriptions during inference. However, the weighted sum method poses challenges, as the speech-driven model exerts significantly stronger correction forces than the text-driven model. This imbalance often generates motions that overly favor speech conditions while suppressing text conditions (Appendix~E). Regarding pseudo labels, the reliance on speech transcripts as supervisory signals creates a domain gap between the transcript distribution and the motion-description distribution, thereby constraining generalization to diverse motion descriptions.

\subsection{Diffusion-based Methods in Motion Generation}
The diffusion model has been recognized as one of the most advanced generative paradigms and has also gained significant attraction in human motion generation. Its applications can be categorized into two primary streams according to the space where the denoising process is implemented. One representative method in the first stream is the Human Motion Diffusion Model (MDM) \cite{tevet2022mdm}, which operates directly in the original motion space to perform denoising \cite{karunratanakul2023guided, alexanderson2023listen, zhu2023tamingspeech, huang2023diffusion, liang2024intergen}. This approach offers several advantages, including high editability and strong controllability, enabling operations such as concatenation and combination within the original motion space during the denoising process. The second stream is exemplified by the Motion Latent Diffusion model (MLD) \cite{chen2023executing}, which conducts denoising in the latent space of a Variational Autoencoder (VAE) \cite{barquero2023belfusion, sampieri2024length, dai2024motionlcm}. Utilizing the VAE latent space improves efficiency in representing complex motion data and significantly reduces the computational load required for diffusion processes, resulting in faster inference speeds. In this work, we select MDM as the foundational model due to its superior editability \cite{shafir2023human,athanasiou2023sinc, petrovich2024multi}.

\subsection{\rev{Compositional Motion Generation}}
\rev{A complementary line of work generates complex motion by \emph{composing} multiple actions or sub-motions. Spatial composition assembles motions of different body parts: SINC \cite{athanasiou2023sinc} combines multiple text-driven actions through body-part masks in a VAE-based, offline manner. Temporal composition instead sequences multiple actions over time and focuses on seamless transitions between them, as in TEACH \cite{athanasiou2022teach}, T2LM \cite{lee2024t2lm}, and FlowMDM \cite{barquero2024seamless}. STMC \cite{petrovich2024multi} unifies both axes, bringing per-denoising-step body-part composition into the diffusion framework with a DiffCollage-style \cite{zhang2023diffcollage} mechanism for temporally overlapping actions. All these methods compose homogeneous, text-only conditions; we instead build on the per-step composition paradigm and extend it to heterogeneous audio, text, and trajectory control, placing \modelname at the intersection of compositional and multi-modal motion generation.}

\section{Method}
\label{sec:method}

In this section, we begin with a brief introduction to the Motion Diffusion Model (MDM) \cite{tevet2022mdm}, which serves as our foundational model (\cref{s_sec:mdm}). Then, we describe our framework's data representations and important model components (\cref{s_sec:data_rep}). Next, we explain our multi-modal decoupled denoising strategy for holistic motion generation (\cref{s_sec:mmdd}). Finally, we address a more complex scenario where the trajectory condition is included and utilize a Large Language Model for motion planning (\cref{s_sec:traj}). An overview of our framework can be found in \cref{fig:framework}.

\definecolor{newblue}{RGB}{72, 116, 203}
\newcommand{\frameworkCaption}{
\textbf{Overview of \modelname.} At each denoising step \( t \), the input conditions and noisy data are fed into their respective denoisers to predict clean motion, which is then diffused for the next iteration. Specifically, the upper-body motion conditioned on speech audio and the lower-body motion conditioned on text description are combined to form the overall body motion. 
The~\textcolor{newblue}{blue} arrows in the figure highlight two key points. One indicates that the denoising process of \( v_0 \) is completed before body motion denoising. The other shows that after the denoising process, the detailed facial and hand movements, and the combined body motion are integrated together to produce the final holistic motion.
}

\begin{figure*}
    \centering

    \includegraphics[width=\linewidth]{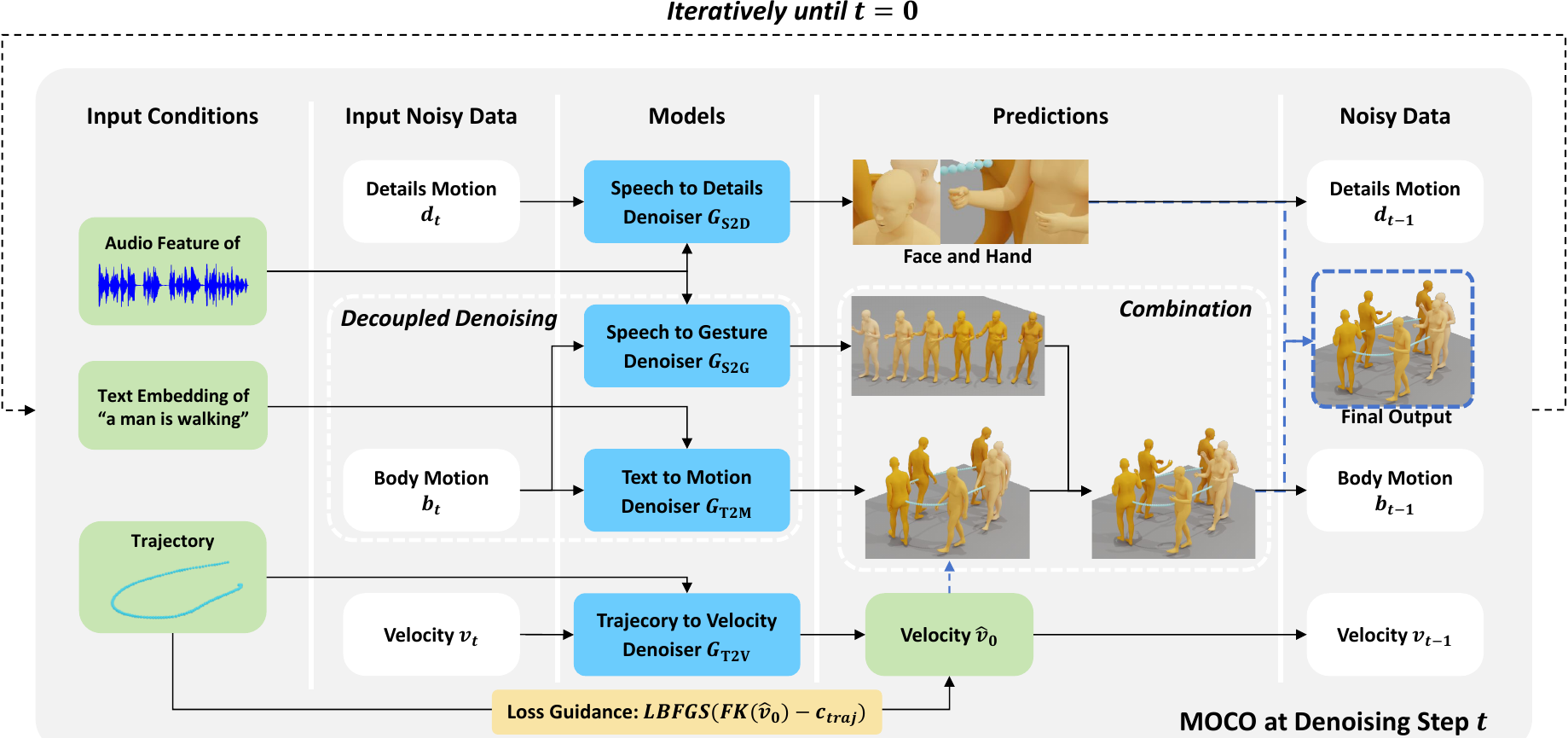}
    \caption{\frameworkCaption}
    \label{fig:framework}
\end{figure*}

\subsection{Preliminary: Motion Diffusion Model}
\label{s_sec:mdm}

MDM is a diffusion-based motion generation model \cite{tevet2022mdm}. The same as other diffusion models, it regards diffusion as a Markov noising process \((x_t)_{t=0}^T\) that starts from a training sample \(x_0\). As the time step \(t\) increases, the distribution of \(x_T\) approaches a standard normal distribution.
Besides, MDM models the conditional distribution \(p(x_0 \mid c)\) by reversing the diffusion process through iterative denoising on \(x_T\). To achieve this, it minimizes  a loss function that penalizes the difference between the original and denoised samples. Moreover, it performs sampling from \(p(x_0 \mid c)\) iteratively, predicting \(x_0\) at each timestep \(t\) and computing \(x_{t-1}\) until $t$ reaches 0.

MDM adopts the classifier-free guidance \cite{ho2022classifier} to adjust its adherence to the conditioning signal \(c\). Specially, its denoiser \(G\) is trained on both conditioned and unconditioned data by randomly setting $c$ as \(\varnothing\), which allows \(G(x_t, t, \varnothing)\) to approximate the unconditional data distribution. During sampling, it adjusts the strength of $c$ using a scaling factor \(s\):
\begin{equation}
G^s\left(x_t, t, c\right) = G\left(x_t, t, \varnothing\right) + s \cdot \left( G\left(x_t, t, c\right) - G\left(x_t, t, \varnothing\right) \right),
\end{equation}
where \(G^s\) denotes the denoiser with the classifier-free guidance.

\subsection{Data Representation and Model Architecture}
\label{s_sec:data_rep}

\paragraph{Data Representation.} Our framework incorporates four data modalities: motion, text, audio, and trajectory. The motion data is represented as $m = \{m^n\}|^{N}_{n=1} \in \mathcal{R}^{N\times 491}$, where $N$ is the number of frames. Specifically, the motion data for each frame is denoted as $m^n = \{b^n, d^n\}$, with $b \in \mathcal{R}^{205}$ representing the body pose \cite{petrovich2024multi},
and $d \in \mathcal{R}^{286}$ capturing detailed facial expression and hand movements. The text embeddings are encoded using the CLIP model \cite{radford2021clip} and are denoted as $c_{text} \in \mathcal{R}^{512}$. Audio features are extracted via the Wav2Vec2 model \cite{baevski2020wav2vec} and represented as $c_{audio} \in \mathcal{R}^{N \times 768}$. Finally, the trajectory data is encoded as $c_{traj} \in \mathcal{R}^{N \times 2}$, representing the position on the XY-plane for each frame.

\paragraph{Model Design.}
Our framework includes four transformer-based denoisers: one for text-to-motion (T2M), one for speech-to-gesture (S2G), one for trajectory-to-velocity (T2V), and one for speech-to-details (S2D) that synthesizes facial expressions and hand poses:
\begin{align}
    &\hat{b}_0 = G_{\text{T2M}}(b_t, t, c_{text}) , \ \ \ \hat{b}_0 = G_{\text{S2G}}(b_t, t, c_{audio}),   \\
    &\hat{v}_0 = G_{\text{T2V}}(v_t, t, c_{traj}) \label{eq:traj}, \ \ \ \hat{d}_0 = G_{\text{S2D}}(d_t, t, c_{audio}).
\end{align}
We denote the sampling with classifier-free guidance for each denoiser as $G^s_{u}$, where $ u \in \{ \text{T2M, S2G, T2V, S2D} \}$.

For the T2M denoiser conditioned on a text embedding $c_{text}$, we follow prior works \cite{tevet2022mdm, chen2023executing} by treating $c_{text}$ as a token and applying self-attention to incorporate the semantic information in $c_{text}$ into the motion generation process. In contrast, the other denoisers are conditioned on sequential data; therefore, we utilize cross-attention to model the relationships between the input sequences and the generated motion. Additionally, the T2M and T2V denoisers are trained on HumanML3D \cite{guo2022humanml3d}, while the S2G and S2D denoisers are trained on the BEAT2 \cite{liu2023emage} dataset. All denoisers adhere to the objective function and diffusion paradigm described in \cref{s_sec:mdm}. The computational complexity of each component is reported in Appendix~F.

\definecolor{newblue}{RGB}{72, 116, 203}
\newcommand{\conditionCaption}{
\textbf{Examples of synchronous and asynchronous conditions.} Synchronous conditions occur when all condition signals are provided within the same time interval. In contrast, asynchronous conditions involve multiple conditions, each corresponding to different time intervals.
}

\begin{figure*}
    \centering

    \includegraphics[width=\linewidth]{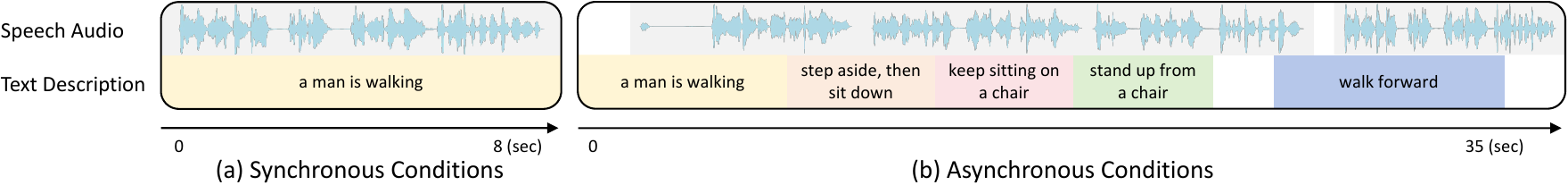}
    \caption{\conditionCaption}
    \label{fig:condition}
\end{figure*}

\subsection{Multi-Modal Decoupled Denoising}
\label{s_sec:mmdd}

In this section, we introduce our multi-modal decoupled denoising strategy for generating motion in scenarios where text and speech audio are provided synchronously—that is, within the same time interval—as shown in \cref{fig:condition} (a). 

Our key observation is that the speech audio naturally guides upper-body gestures (including head and arm poses), while text descriptions mainly influence lower-body movements (including spine and leg poses) like walking or shifting stance. In a diffusion context, this implies that the joint conditional probability \( p(b_{t-1} \mid c_{\text{text}}, c_{\text{audio}}, b_t) \) can be approximated by the product of two probabilities: \( p(b_{t-1, \text{lower}} \mid c_{\text{text}}, b_{t}) \) and \( p(b_{t-1, \text{upper}} \mid c_{\text{audio}}, b_{t}) \). Mathematically, this can be represented as follows:
\begin{align}
    p(b_{t-1} \mid c_{\text{text}}, c_{\text{audio}}, b_t) \approx 
    p(b_{t-1, \text{lower}} \mid c_{\text{text}}, b_{t}) \cdot p(b_{t-1, \text{upper}} \mid c_{\text{audio}}, b_{t}).
    \label{eq:ta_0}
\end{align}
Here, \( b_t \) represents the motion at denoising step \( t \), which is composed of the upper-body motion \( b_{t, \text{upper}} \) and the lower-body motion \( b_{t, \text{lower}} \). A detailed derivation of \cref{eq:ta_0} is provided in Appendix~B.

Based on this observation, we can decouple the multi-modal controlled denoising process into distinct streams. The text-driven stream $p(b_{t-1} \mid c_{\text{text}}, b_{t})$ can be formulated as follows:
\begin{align}
    \hat{b}^{\text{text}}_0 = G^s_{\text{T2M}}(b_t, t, c_{\text{text}})   , \ \ \ b^{\text{text}}_{t-1} = \sqrt{\alpha_{t-1}} \, \hat{b}^{\text{text}}_0 + \sqrt{1 - \alpha_{t-1}} \, \epsilon,  
\end{align}
where \( \alpha_{t} = \prod_{s=1}^{t} (1 - \beta_s) \) represents the cumulative product of \( (1 - \beta_s) \) up to timestep \( t \), and \( \beta_s \) is the variance schedule controlling the amount of noise added at each timestep. The variable \( \epsilon \sim \mathcal{N}(0, \mathbf{I}) \) is Gaussian noise sampled from a standard normal distribution \cite{tevet2022mdm}. Similarly, the audio-driven stream $p(b_{t-1} \mid c_{\text{audio}}, b_{t})$ is defined as:
\begin{align}
    \hat{b}^{\text{audio}}_0 = G^s_{\text{S2G}}(b_t, t, c_{\text{audio}})  , \ \ \ b^{\text{audio}}_{t-1} = \sqrt{\alpha_{t-1}} \, \hat{b}^{\text{audio}}_0 + \sqrt{1 - \alpha_{t-1}} \, \epsilon.  
\end{align}
Finally, the body motion $b_{t-1}$ at step $t$ is computed as follows:
\begin{equation}
    b_{t-1} = I \odot b^{\text{text}}_{t-1} + (1 - I) \odot b^{\text{audio}}_{t-1},  
\end{equation}
where \( I \in \mathbb{R}^{205} \) is a binary vector with entries set to 1 for the lower body and 0 for the upper body; \( \odot \) denotes element-wise multiplication.

The decoupled denoising allows each body part's motion to be precisely guided by its corresponding input condition, ensuring high fidelity to the control signals. Furthermore, each stream adjusts its output based on the current overall motion, promoting coordination among the body parts. As this process continues, the motions generated for different body parts become increasingly synchronized, resulting in natural and coherent full-body movements.

\subsection{Trajectory and Motion Planning}
\label{s_sec:traj}
The above subsection handles the multi-modal decoupled denoising for synchronous conditions. We now extend our \modelname framework to tackle more complex, challenging yet more realistic scenarios, such as incorporating trajectory control and leveraging Large Language Model (LLM) for motion planning.

\paragraph{Trajectory Control.}
Following the approach of Petrovich \etal \cite{petrovich2024multi}, we represent the global transition of body pose using the velocity vector \( v = [\dot{r}_x, \dot{r}_y, \dot{\theta}] \), where \( \dot{r}_x \) and \( \dot{r}_y \) are the linear velocities of the pelvis in the \( x \) and \( y \) directions, respectively, and \( \dot{\theta} \) is the angular velocity about the body's vertical (Z) axis. Given the trajectory data \( c_{\text{traj}} \), we first predict \( \hat{v}_0 \) using \cref{eq:traj}. To enhance prediction accuracy, we incorporate loss guidance into our method.
During each denoising step for predicting the velocity vector, we compute \( \hat{v}_0 \) using \cref{eq:traj} and apply loss guidance as follows:
\begin{equation}
L_{\text{guidance}} = FK(\hat{v}_0) - c_{\text{traj}},  
\label{eq:guidance}
\end{equation}
where \( FK \) represents the differentiable Forward Kinematics function \cite{wang2023intercontrol} that converts linear and angular velocities into the trajectory. We follow the methodology of InterControl \cite{wang2023intercontrol} and optimize \( L_{\text{guidance}} \) with respect to \( \hat{v}_0 \) using the second-order LBFGS optimizer \cite{liu1989bfgs}. This optimization ensures that the predicted global transitions closely match the provided trajectory data.

Once \( \hat{v}_0 \) is predicted, we substitute the velocity component in \( \hat{b}_0 \) with \( \hat{v}_0 \) during each iteration of its generation. This substitution guides the generation process to adapt the remaining elements of \( \hat{b}_0 \) to align with \( \hat{v}_0 \), thereby ensuring consistency with the provided trajectory data.


\begin{table}[h]

\scriptsize
\caption{Example of an LLM-generated motion timeline.}
\label{tab:timeline_example}
    \centering
    \begin{tabular*}{\linewidth}{@{\extracolsep{\fill}}lcccc}
    \toprule
Condition & Start Time & End Time & Body Part \\ \hline
$c_1$\ =\ kneel down & $f^{s}_1$ & $f^{e}_1$ & spine, legs \\
$c_2$\ =\ stand up from ground & $f^{s}_2$ & $f^{e}_2$ & spine, legs \\
$c_3$\ =\ run forward & $f^{s}_3$ & $f^{e}_3$ & spine, legs \\
$c_4$\ =\ audio clip 1 & $f^{s}_4$ & $f^{e}_4$ & head, arms \\
$c_5$\ =\ audio clip 2 & $f^{s}_5$ & $f^{e}_5$ & head, arms \\ \bottomrule
    \end{tabular*}

\end{table}

\paragraph{Motion Planning.}
In real-world applications, models could face audio conditions paired with complex text descriptions, which can challenge their ability to handle intricate semantics. One approach to address this challenge is decomposing complex textual descriptions into basic motion units and generating them separately \cite{sun2024prompt}. However, this decomposition often introduces more complex conditions—such as asynchronous conditions, as illustrated in \cref{fig:condition}.

To address these challenges, we propose an LLM-based motion planning pipeline. Our pipeline begins by using an LLM to generate a ``motion timeline'' from the input text and audio conditions. Particularly, we carefully design a prompt that instructs the LLM to decompose complex text into elementary motion units, insert appropriate pose transitions, and specify the start and end times and the body parts involved for each condition. More details are provided in Appendix~C. For example, when given the text condition “kneel down then walk forward” along with two audio clips, the LLM produces a motion timeline comprising multiple intervals, as detailed in \cref{tab:timeline_example}.

Notably, the transition condition \(c_2\) is automatically generated by the LLM, which ensures a smooth stance changes and avoids abrupt movements. Users, however, have the flexibility to customize the timeline. We then adopt the strategy of STMC \cite{petrovich2024multi}, generating body masks according to a predefined timeline. At each denoising step \(t\), the body pose across the entire timeline is generated as follows:
\begin{equation}
    \hat{b}_0 = \sum_{j=1}^{J} I_j \odot G^s_j\left(b_{t, f^{s}_j:f^{e}_j}, t, c_j\right),
\end{equation}
where \(I_j\) represents a binary mask corresponding to the motion generated by the \(j\)-th condition, and \(G^s_j \in \{G^s_{\text{T2M}}, G^s_{\text{S2G}}\}\) denotes the denoiser utilized for the \(j\)-th condition. The operator \(\odot\) stands for element-wise multiplication.

Similarly, the facial and hand movements at step \(t\) are computed by:
\begin{equation}
\hat{d}_0 = \sum_{j=1}^{J} G_{\text{S2D}}\left(d_{t, f^{s}_j:f^{e}_j}, t, c_j\right).
\end{equation}
Since the text-to-motion dataset, HumanML3D, lacks facial and hand movement data, the denoiser \( G_{\text{S2D}} \) is only trained on the speech-to-gesture dataset. Therefore, in intervals without audio clips, we maintain the model's temporal continuity by setting the conditional input \( c_j \) to the empty condition \( \varnothing \). Finally, we employ DiffCollage \cite{zhang2023diffcollage} to ensure smoother transitions at the interval boundaries.
\section{Experiments}
\label{sec:exp}

\subsection{Datasets}
\label{sec:dataset}

\paragraph{Task-Specific Datasets.}
The \underline{HumanML3D} dataset is a large \textbf{Text-to-Motion} dataset created by amalgamating motion sequences from the HumanAct12 and AMASS datasets \cite{guo2022humanml3d}. It consists of 14,616 motions and 44,970 descriptions composed of 5,371 distinct words, totaling 28.59 hours of motion data. To align the data representation—specifically, to use SMPL-X parameters for representing joint rotations—we utilize only the AMASS portion of HumanML3D because it has an official SMPL-X version. The \underline{BEAT2} dataset is a large-scale \textbf{Speech-to-Gesture} dataset specifically designed for research in speech-to-gesture generation \cite{liu2023emage}. It contains synchronized recordings of speech audio and corresponding 3D motion capture data of human gestures, totaling 60 hours of data from 25 speakers. In addition to audio and motion data, the dataset includes valuable annotations such as text transcriptions and emotional states. Here, we select speakers with IDs lower than 10, resulting in a total of 10.15 hours of training data.

\paragraph{Multi-Modal Benchmark.} We create a multi-modal benchmark of 1,000 test clips to evaluate our proposed task effectively.
Each test clip is automatically constructed and contains two text descriptions and two audio clips. To create these clips, we manually collect 40 texts focusing on lower-body movements that commonly occur during speech delivery or conversation. We then split the audio from the BEAT2 test set into clips using a Voice Activity Detector (VAD). To serve as ground truth for computing evaluation metrics (\cref{sec:metrics}), we selected motion samples from AMASS and BEAT2 corresponding to each text and audio clip. This structured construction enables controlled and repeatable comparisons under concurrent multi-modal conditions and isolates the challenge of resolving conflicting cross-modal signals. Accordingly, the benchmark is intended to probe compositional consistency rather than broad coverage of open-ended semantics. More details about the benchmark are provided in Appendix~H.

\subsection{Metrics} 
\label{sec:metrics}
We evaluate our method using two categories of metrics: text-to-motion and speech-to-gesture \cite{guo2022humanml3d, liu2023emage, petrovich2024multi}. For text-to-motion, \textit{FID+} assesses realism by measuring the distribution difference between real and generated motions using five random 5-second clips per test sample. \textit{R1} evaluates alignment by recording the frequency of correct text prompts appearing in the top-1 retrieved text. \textit{M2T} (motion-to-text) and \textit{M2M} (motion-to-motion) measure alignment through cosine similarity between embeddings of generated motions and ground truth texts or motions. In the speech-to-gesture category, \textit{FID-A} similarly measures the realism of motion generated based on speech audio. \textit{Beat Consistency (BC)} evaluates how well gestures synchronize with the rhythm and beats of the speech, while \textit{L1 Diversity (L1Div)} quantifies gesture diversity by calculating the average L1 distance between multiple gesture clips.
This comprehensive set of metrics thoroughly evaluates our method across key dimensions.

\subsection{Comparison with Baselines}
\label{sec:baseline}
\paragraph{Baseline Setting.}
To effectively evaluate the performance of our method, we propose a series of baselines inspired by reasonable settings and prior works. These include:
\textit{Weighted Sum}, a method that follows \cite{yang2024freetalker} by combining the predictions of text- and audio-conditioned models through weighted sum; \textit{Pseudo-Text}, a method that follows \cite{ling2023mcm} by using pseudo text descriptions of a speaker's speech as the text condition during training; and \textit{SynTalker} \cite{chen2024enabling}, an open-source multi-modal motion generation method performing weighted sum in body-part-specific latent space. Notably, because these baselines cannot handle asynchronous conditions, we align the text time intervals with the audio time intervals to construct synchronous conditions for all baseline methods. The results are reported in \cref{tab:baseline}.

\begin{table}[t]
\centering
\caption{Comparison with baselines. We highlight the best performance in \textbf{bold} and use \underline{underlining} to indicate the second-best performance.}
\label{tab:baseline}

\begin{tabular*}{\linewidth}{@{\extracolsep{\fill}}l|cccc|ccc} 
\toprule

                    & \multicolumn{4}{c|}{Text2Motion} & \multicolumn{3}{c}{Speech2Gesture}    
                    \\
                    & FID+ $\downarrow$  & R1 $\uparrow$  & M2T $\uparrow$ & M2M $\uparrow$ & FID-A $\downarrow$   & BC $\uparrow$      & L1div $\uparrow$  \\ \midrule
GT (Ground Truth)                 & 0.000 & 40.0  & 0.781 & 1.000 & - & - & - \\
Weighted Sum \cite{yang2024freetalker}  & 1.335             & 6.8               & 0.546         & 0.537 & \textbf{2.17} & 2.20 & 4.08    \\
Pseudo-Text      \cite{ling2023mcm}     & 1.593             & 2.2               & 0.511         & 0.503 & \underline{2.22} & 2.55 & 6.43  \\ 
SynTalker      \cite{chen2024enabling}  & \underline{0.985}             & \underline{9.8}               & \underline{0.601}         & \underline{0.603} & 6.60 & \textbf{2.95} & \textbf{9.12}   \\ \
\textbf{\modelname}                     & \textbf{0.862} & \textbf{24.6}       & \textbf{0.649}  & \textbf{0.639} & 3.83 & \underline{2.72} & \underline{8.62}   \\ 
\bottomrule
\end{tabular*}
\end{table}

\paragraph{Result Analysis.}
As shown in the table, our proposed method, \modelname, exhibits robust performance across both sets of metrics, delivering competitive results in both text-to-motion and speech-to-gesture tasks simultaneously. This underscores the effectiveness of \modelname in generating condition-aligned motions when multi-modal conditions are provided concurrently.

\textit{Weighted Sum} and \textit{Pseudo-Text} achieves good results on speech-to-gesture metrics but poor performance on text-to-motion metrics, indicating their limited ability to handle multi-modal data. We explain this further in Appendix~E.

\textit{SynTalker} also notices the importance of separating different body parts, resulting in an improvement in text-to-motion metrics compared to \textit{Weighted Sum} and \textit{Pseudo-Text}. However, its insufficient decoupling of conditions and suboptimal joint latent space—negatively impacted by repetitive patterns in the speech-to-gesture data—results in limited expressiveness and unstable movements. This is reflected in its less competitive performance in R1 and FID-A.

\rev{Beyond the baselines above, STMC~\cite{petrovich2024multi} is particularly relevant, as its per-step composition paradigm inspired our multi-modal decoupled denoising. Since STMC is text-only and does not natively accept speech audio, we adapt it to our concurrent setting and conduct a dedicated comparison, including a user study (Appendix~G.1 and Appendix~G.5). There, STMC attains competitive text-to-motion scores, yet \modelname{} produces more natural and better-synchronized motion, as confirmed by quantitative metrics, additional naturalness metrics, and the subjective user study.}

\subsection{Ablation Study}
\label{sec:ablation}
To assess the impact of key designs within \modelname, we conduct an ablation study presented in \cref{tab:ablation}. This study systematically examines the effects of body masking (\textit{Body Mask}), weight sharing (\textit{Share Weight}), and combination at each step (\textit{Per-Step Comb.})
on the model's performance across text-to-motion and speech-to-gesture metrics. In addition, we introduce the \textit{Transition Smoothness Ratio (TSR)} to measure temporal coherence at segment boundaries, defined as the ratio between the mean frame-to-frame speed within transition windows and the mean speed in non-transition regions.

\paragraph{Body Masking.} In Variants 1 and 2, we test our hypothesis that speech audio guides upper-body motion (head and arms) while text descriptions influence lower-body movements (spine and legs). In Variant 1, we expand the body mask to include the spine along with the head and arms (\textit{Body Mask} = head, arms, spine). This modification results in a worse FID+ and a slight decrease in R1, indicating a decline in text-to-motion performance. Moreover, it does not produce significant improvements in speech-to-gesture metrics, suggesting that including the spine in the body mask fails to enhance gesture generation and instead compromises text-driven motion performance.

Variant 2 further adjusts the body mask to include the legs and spine (\textit{Body Mask} = legs, spine), which significantly degrades text-to-motion metrics and Beat Consistency. This mainly stems from a mismatch: the texts in our multimodal benchmark describe diverse actions such as ``walk,'' ``sit,'' and ``turn right,'' whereas the speech-to-gesture data are mostly standing gestures. Controlling lower-body motion with audio makes it hard to align with these texts, while controlling upper-body motion with text complicates beat alignment. Although Variant 2 yields a notable improvement in FID-A—reflecting a bias in the speech-to-gesture data toward standing-in-place motions—its overall performance still degrades.

In contrast, our original method (\textit{Body Mask} = head, arms) effectively balances the influences of both text and audio inputs. By assigning the upper body to be guided by audio and the lower body by text, we achieve superior results across both text-to-motion and speech-to-gesture metrics. This demonstrates the advantage of our approach in producing coherent and contextually appropriate motions that align well with the provided conditions.

\paragraph{Weight Sharing.} In Variant 3, we enable weight sharing (\textit{Share Weight} = \checkmark), following previous multi-modal methods \cite{ling2023mcm, yang2024freetalker, chen2024enabling}, while keeping the body mask and transition method unchanged. Compared to the full \modelname model (without weight sharing), enabling weight sharing results in worse performance across several metrics, including R1 and FID-A. This decline suggests that sharing weights between modalities may limit the model's ability to capture modality-specific nuances, thereby reducing its effectiveness in generating accurate and realistic motions for both text-to-motion and speech-to-gesture tasks.

\paragraph{Combination at Each Step.} 
Variant 4 performs body-part combination only at the final denoising step $t=0$ (\textit{Per-Step Comp.} = \ding{55}). 
Lacking iterative coordination to aggregate holistic motion information, this baseline 
exhibits temporal discontinuities (reflected by high TSR) and reduced body coherence. 
A user study (\cref{sec:user}) and qualitative results in the \textbf{supplementary videos} 
further confirm that \modelname produces more natural and synchronized motions.

\begin{table}[t]
\centering
\scriptsize

\caption{Ablation study on key designs within \modelname. We highlight the best performance in \textbf{bold} and use \underline{underlining} to indicate the second-best performance achieved by our method.}
\label{tab:ablation}
\begin{tabular*}{\linewidth}{@{\extracolsep{\fill}}c|ccc|ccc|ccc|c} 
\toprule

Method & \multirow{2}{*}{\shortstack{Share\\Weight}} & \multirow{2}{*}{\shortstack{Body Mask \\ $1-I$}}  & \multirow{2}{*}{\shortstack{Per-Step\\Comb.}}
&\multicolumn{3}{c|}{Text2Motion} & \multicolumn{3}{c|}{Speech2Gesture}   
& \multirow{2}{*}{TSR $\downarrow$}  
\\
& & & & FID+ $\downarrow$  & R1 $\uparrow$  & M2T $\uparrow$ & FID-A $\downarrow$   & BC $\uparrow$      & L1div $\uparrow$ \\ \midrule
GT &-&-  & -                       & 0.000 & 40.0 & 0.781 & - & - & -  & -  \\ 
\textbf{\modelname} &\ding{55} & head, arms  &\checkmark  & \underline{0.862} & \textbf{24.6} & \underline{0.649} & \underline{3.83} & \underline{2.72} & \underline{8.62}  & \underline{0.90}  \\
Variant 1 &\ding{55} & head, arms, spine   &\checkmark  & 0.921 & 22.4 & 0.634& 3.86 & \textbf{2.81} & 8.35 & 0.90  \\
Variant 2 &\ding{55} & spine, legs    &\checkmark & 1.234 & 7.9  & 0.554& \textbf{2.75} & 2.19 & 5.11 & 0.95 \\
Variant 3 &\checkmark & head, arms &\checkmark & 0.866 & 22.1 & \textbf{0.656} & 4.24 & 2.68 & \textbf{8.95}    & \textbf{0.85}      \\ 
Variant 4 &\ding{55} & head, arms & \ding{55} & \textbf{0.832}      & {24.4}  & {0.645}& 3.99 & 2.27 & 8.60 & 1.17  \\

\bottomrule
\end{tabular*}

\end{table}

\subsection{Qualitative Analysis}
\label{sec:qual}

\newcommand{\qualitativeCaption}{
\textbf{Qualitative results.} 
We visualize four samples generated by \modelname. Darker colors represent later points in time. The results demonstrate that \modelname is capable of generating coherent and realistic motions that highly align with the given multi-modal control signals. Figures (a), (b), and (d) present natural speech gestures coordinated with various lower-body movements as specified by the text inputs, such as jogging, walking in a circle, turning right, running, and so on. Figure (c) displays natural  speech movements while sitting down. Figure (d) reveals a limitation of \modelname. When standing up or sitting down, the foot should remain stationary. However, the foot highlighted in the red box slides, leading to unrealistic results.
}

\begin{figure*}
    \centering

    \includegraphics[width=\linewidth]{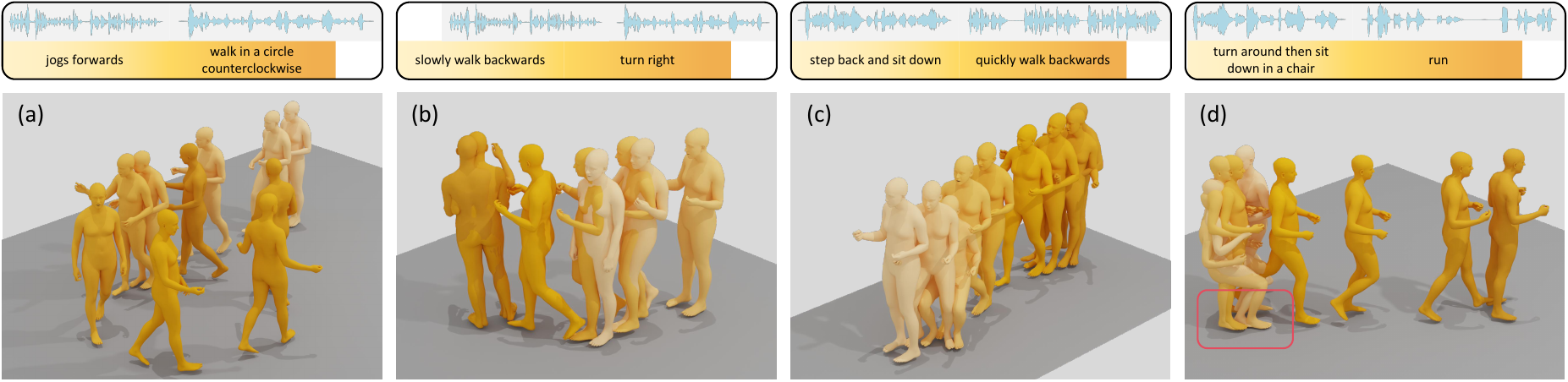}
    \caption{\qualitativeCaption}
    \label{fig:qualitative}
\end{figure*}

To clearly illustrate the overall performance of \modelname, we visualize four generated samples and their corresponding conditions in \cref{fig:qualitative}. The lighter color of the mesh and the text description background indicates the start of each sequence, while the darker color indicates the end. These results exhibit natural speech-driven upper-body gestures synchronized with diverse lower-body motions such as jogging, walking, and sitting, demonstrating \modelnames ability to generate coherent and realistic motions that closely align with the given multimodal signals. A detailed analysis of these examples is provided in the figure caption. Additional qualitative results, including comparisons with baselines, are included in the supplementary materials.

\begin{table}[ht]
\centering
\scriptsize
\caption{User study results comparing MOCO against baseline methods. }
\label{tab:user}
\begin{tabular*}{\linewidth}{@{\extracolsep{\fill}}lcccccc}
\toprule
\multirow{2}{*}{Baseline} & \multicolumn{3}{c}{Better Text Following (\%)} & \multicolumn{3}{c}{Better Beat Synchronization (\%)} \\ \cmidrule(lr){2-4} \cmidrule(lr){5-7}
& Neither & Baseline & \modelname (Ours) & Neither & Baseline & \modelname (Ours) \\ \midrule
Pseudo-Text & 1.0 & 0.0 & \textbf{99.0} & 13.0 & 12.5 & \textbf{74.5} \\
Weighted Sum & 0.0 & 0.0 & \textbf{100.0} & 13.5 & 3.5 & \textbf{83.0} \\ \midrule\midrule
\multirow{2}{*}{Baseline} & \multicolumn{3}{c}{Better Body Coherence (\%)} & \multicolumn{3}{c}{Better Temporal Fluidity (\%)} \\ \cmidrule(lr){2-4} \cmidrule(lr){5-7}
& Neither & Baseline & \modelname (Ours) & Neither & Baseline & \modelname (Ours) \\ \midrule
\multirow{2}{*}{\shortstack{Combine Once \\ (Variant 4)}} & \multirow{2}{*}{12.0} & \multirow{2}{*}{17.0} & \multirow{2}{*}{\textbf{71.0}} & \multirow{2}{*}{12.6} & \multirow{2}{*}{42.6} & \multirow{2}{*}{\textbf{44.8}} \\
 &  &  &  &  &  &  \\
\bottomrule
\end{tabular*}
\end{table}

\subsection{User Study}
\label{sec:user}

\cref{tab:user} presents the results of a user study comparing \modelname with three methods. Specifically, we evaluate \modelname against \textit{Pseudo-Text} and \textit{Weighted Sum}, introduced in \cref{sec:baseline}, and \textit{Combine Once} (Variant 4 in \cref{sec:ablation}), to assess overall performance in text following and audio beat synchronization. Additional experimental details are provided in Appendix~G.4.

As shown in the table, \modelname achieves significant advantages over both \textit{Pseudo-Text} and \textit{Weighted Sum}, demonstrating the effectiveness of our decoupled denoising strategy in generating motion aligned with multi-modal conditions. Furthermore, when compared to \textit{Combine Once}, our method was rated significantly higher in both body coherence and temporal fluidity. This indicates that \modelname does more than merely combine different body parts controlled by separate conditions; it ensures that each body part aligns with its corresponding condition while enhancing coordination among all body parts.
To further strengthen these findings, we additionally conduct a larger-scale pairwise preference study with 51 participants comparing \modelname{} against STMC, where \modelname{} is preferred with statistical significance on audio synchronization and naturalness ($p<0.05$); full protocol and results are reported in the supplementary material.

\section{Discussion and Limitations}
\label{sec:dis}
MOCO achieves effective multi-modal motion generation, but its fixed body-part assignment can be suboptimal when control signals conflict. For instance, when text descriptions specify arm movements (\eg, “wave hands”), this rigid separation leaves the upper body solely governed by audio, preventing the model from following text instructions involving arm motions. To mitigate this limitation, MOCO allows users to configure body-part control through the motion timeline (see Appendix~D), enabling text to drive upper-body motion when needed.

Our multi-modal benchmark is also deliberately focused on compositional consistency under controlled settings rather than open-ended semantic coverage, enabling precise and repeatable evaluation of cross-modal integration but not claiming generalization to arbitrary text–audio pairs. Future work may explore adaptive body-part assignment or residual blending that dynamically handles spatially overlapping actions, moving beyond fixed spatial rules toward context-aware fusion while keeping the interpretability of decoupled generation.

\section{Conclusion}
\label{sec:con}

This study presents \textbf{\modelname}, a novel diffusion-based framework to generate realistic and coherent holistic body motions from multi-modal control signals, including text descriptions, speech audio, and trajectory data. Our key innovation lies in a decoupled denoising process where, during each denoising step, the model independently generates motions for each modality and assembles them according to predefined spatial rules. This approach ensures that the generated motion is closely aligned with each condition while producing realistic and coherent whole-body movements. Experimental results demonstrate that our approach delivers state-of-the-art performance both qualitatively and quantitatively, advancing the field of multi-modal controlled motion generation.

\medskip

\noindent
{\qheading{Acknowledgments.} This work was supported by the National Natural Science Foundation of China under Grants 62476099 and 62076101, Guangdong Basic and Applied Basic Research Foundation under Grants 2024B1515020082 and 2023A1515010007, and the TCL Young Scholars Program.}


%
%
\bibliographystyle{splncs04}
\bibliography{Bibs/sample-base}

\clearpage
\appendix
\renewcommand{\theHsection}{appendix.\Alph{section}}
\section{Body-Part Motion Statistics}

\label{sup:support}

A core design principle of \textsc{MOCO} is the spatial decomposition of 
multi-modal conditioning: text descriptions govern lower-body motion, while 
speech audio drives upper-body dynamics. This appendix provides a quantitative 
characterization of the motion distributions associated with each conditioning 
modality within the scope of our target task, offering an empirical basis for 
understanding why this decomposition is well-suited to the concurrent 
text-and-speech control scenario.

\subsection*{Task-Scoped Text Conditions and Motivation for Generated Samples}

Our target application involves concurrent control by both speech audio and 
text descriptions. In this setting, text conditions are expected to specify 
body-level actions that are \textit{spatially complementary} to co-speech 
gestures---that is, actions primarily activating the lower body without 
conflicting with concurrent upper-body gesture generation. We therefore focus 
our analysis on a curated set of 40 text descriptions centered on locomotion 
and postural transitions (see \aref{sup:benchmark}), which represent the intended
operating domain of the text conditioning signal in \textsc{MOCO}. These 
descriptions are representative of the text conditions under which \textsc{MOCO} 
is designed to operate, and do not aim to characterize the full diversity of 
text-driven motion.

To obtain the motion distribution for these curated descriptions, we generate 
1,000 sequences using a pretrained text-to-motion model (MDM~\cite{tevet2022mdm}) 
trained on HumanML3D~\cite{guo2022humanml3d}, as these specific 
lower-body--oriented descriptions do not have corresponding ground-truth 
annotations in existing datasets. This approach allows us to directly estimate 
the body-part activation patterns associated with the text conditions employed 
in our multi-modal benchmark.

\subsection*{Setup}

Both data sources are represented using the unified 205-dimensional SMPL-X 
body parameterization. Body parts are defined via fixed joint indices, grouped 
into five regions: left arm, right arm, head, spine, and legs. For each motion 
sequence, we compute the temporal variance of each body-part region as a proxy 
for motion activity, and normalize across all five parts so that they sum to 
100\% per sequence. We further aggregate upper-body parts (head, left arm, 
right arm) and lower-body parts (spine, legs), and define the lower-body ratio 
as:
\begin{equation}
    r_{\text{lower}} = 
    \frac{\text{Var}_{\text{lower}}}{\text{Var}_{\text{upper}} + 
    \text{Var}_{\text{lower}}},
\end{equation}
where $\text{Var}_{\text{upper}}$ and $\text{Var}_{\text{lower}}$ denote the 
aggregated temporal variance of upper- and lower-body joints, respectively. 
This normalization eliminates confounds from sequence length and global motion 
scale, enabling a fair comparison across datasets.

Statistics are computed over two sources:
\begin{itemize}
    \item \textbf{Generated (text-driven):} 1,000 sequences produced by a 
    pretrained text-to-motion diffusion model conditioned on our curated set 
    of 40 locomotion- and posture-oriented descriptions.
    \item \textbf{BEAT2 (speech-driven):} 5,000 sequences sampled from the 
    BEAT2 training set~\cite{liu2023emage}.
\end{itemize}

\subsection*{Results}

Quantitative results are reported in \cref{tab:body_part_stats}. Under our task-scoped text conditions, the generated motions exhibit a lower-body variance share of 62.0\%, indicating that locomotion- and posture-oriented descriptions predominantly activate the legs and spine. In contrast, BEAT2 speech-gesture sequences are strongly upper-body dominant, with 85.1\% of motion variance concentrated in the head and arms—consistent with the expressive, gesture-rich nature of co-speech motion. These complementary distributions provide empirical support for our proposed spatial decomposition.

\begin{table}[h]
\centering
\caption{
    Body-part motion statistics for text-driven generated sequences and BEAT2 speech-gesture data. Temporal variance is computed per body-part region and normalized within each sequence.
}
\label{tab:body_part_stats}
\begin{tabular}{lcccc}
\toprule
\multirow{2}{*}{\textbf{Source}} & 
\multirow{2}{*}{\shortstack{Upper \\ MeanVar}} & 
\multirow{2}{*}{\shortstack{Lower \\ MeanVar}} & 
\multirow{2}{*}{Lower / Upper} & 
\multirow{2}{*}{\shortstack{Lower-Body \\ Ratio}} \\
 &  &  &  &  \\
\midrule
Generated (text-driven) & 0.00387 & 0.00663 & 1.71$\times$ & 62.0\% \\
BEAT2 (speech-driven)   & 0.00587 & 0.00164 & 0.28$\times$ & 14.9\% \\
\bottomrule
\end{tabular}
\end{table}

\section{Theoretical Analysis for Decoupled Denoising}
\label{sup:theoretical}

Our proposed \modelname relies on the assumption that the joint conditional probability \( p(b_{t-1} \mid c_{\text{text}}, c_{\text{audio}}, b_t) \) can be approximated by \( p(b_{t-1, \text{lower}} \mid c_{\text{text}}, b_{t}) \cdot p(b_{t-1, \text{upper}} \mid c_{\text{audio}}, b_{t}) \), expressed as:
\begin{align}
    & p(b_{t-1} \mid c_{\text{text}}, c_{\text{audio}}, b_t) \notag \\ 
    & \approx 
    p(b_{t-1, \text{lower}} \mid c_{\text{text}}, b_{t}) \cdot p(b_{t-1, \text{upper}} \mid c_{\text{audio}}, b_{t}),
    \label{eq:ta_0_sup}
\end{align}
where \( b_t \) denotes the motion at denoising step \( t \), composed of upper-body motion \( b_{t, \text{upper}} \) and lower-body motion \( b_{t, \text{lower}} \).

\begin{align}
& p(b_{t-1} \mid c_{\text{text}}, c_{\text{audio}}, b_t) \notag 
\\ & = p(b_{t-1, \text{lower}}, b_{t-1, \text{upper}} \mid c_{\text{all}}), \text{ where } c_{\text{all}} = \{c_{\text{text}}, c_{\text{audio}}, b_{t}\} \nonumber\\
&= p(b_{t-1, \text{lower}} \mid c_{\text{all}}) \cdot p(b_{t-1, \text{upper}} \mid c_{\text{all}}, b_{t-1, \text{lower}}) \label{eq1}\\
&\approx p(b_{t-1, \text{lower}} \mid c_{\text{all}}) \cdot p(b_{t-1, \text{upper}} \mid c_{\text{all}}) \label{approx1}\\
&\approx p(b_{t-1, \text{lower}} \mid c_{\text{all}} \setminus \{c_{\text{audio}}\}) \cdot p(b_{t-1, \text{upper}} \mid c_{\text{all}} \setminus \{c_{\text{text}}\}) \label{approx2} \\
&= p(b_{t-1, \text{lower}} \mid c_{\text{text}}, b_{t}) \cdot p(b_{t-1, \text{upper}} \mid c_{\text{audio}}, b_{t}). \nonumber
\end{align}

The first approximation occurs in the transition from \cref{eq1} to \cref{approx1}. Here, we approximate:
\begin{align}
&p(b_{t-1, \text{upper}} \mid c_{\text{all}}, b_{t-1, \text{lower}}) \notag \\ 
&=p(b_{t-1, \text{upper}} \mid c_{\text{text}}, c_{\text{audio}}, b_{t, \text{upper}}, b_{t, \text{lower}}, b_{t-1, \text{lower}}) \nonumber \\
&\approx p(b_{t-1, \text{upper}} \mid c_{\text{text}}, c_{\text{audio}}, b_{t, \text{upper}}, b_{t, \text{lower}}) \nonumber \\
&= p(b_{t-1, \text{upper}} \mid c_{\text{all}}). \nonumber
\end{align}
This approximation assumes that \( b_t \) already encapsulates sufficient information about \( b_{t-1} \), allowing us to neglect the influence of \( b_{t-1, \text{lower}} \) when estimating \( b_{t-1, \text{upper}} \). This simplification is justified by the proximity of the diffusion steps and the strong correlation between the states at steps \( t \) and \( t-1 \).

The second approximation occurs in the transition from \cref{approx1} to \cref{approx2}, where we decouple modality-specific influences:
\[
\begin{aligned}
p(b_{t-1, \text{lower}} \mid c_{\text{all}}) &\approx p(b_{t-1, \text{lower}} \mid c_{\text{all}} \setminus \{c_{\text{audio}}\}), \\
p(b_{t-1, \text{upper}} \mid c_{\text{all}}) &\approx p(b_{t-1, \text{upper}} \mid c_{\text{all}} \setminus \{c_{\text{text}}\}).
\end{aligned}
\]
This approximation leverages the observation that text input (\( c_{\text{text}} \)) primarily influences lower-body movements (\eg, walking or shifting stance), while audio input (\( c_{\text{audio}} \)) predominantly affects upper-body movements (\eg, gestures or facial expressions). By excluding \( c_{\text{audio}} \) from the conditioning set for \( b_{t-1, \text{lower}} \) and \( c_{\text{text}} \) for \( b_{t-1, \text{upper}} \), we ensure the conditioning focuses on the most relevant modality for each body part.

\section{Leveraging LLM for Motion Planning}
\label{sup:llm}

In summary, our prompt instructing the LLM to generate a motion timeline consists of several key components:
\begin{enumerate}
\item Purpose: Clearly state the objective of generating a motion timeline based on text and audio conditions.
\item Output Formats: Specify the formats that the LLM should adhere to when producing output.
\item Guidelines for Generating Timelines: Outline several essential rules that the LLM must follow.
\item Examples of Timelines: Provide a pair of example timelines, showcasing both an effective and a less effective version.
\end{enumerate}

Additionally, to enhance the precision and completeness of the LLM's reasoning, we append the phrase ``please reason step by step'' at the end of the dialogue.

To illustrate this process more clearly, we provide a complete dialogue record with GPT-4o mini in \texttt{motion-planning-dialogue.pdf}. In this dialogue, the input conditions are processed by the LLM as follows:

\begin{table}[h]
    \centering
    \begin{adjustbox}{center, scale = 1}
    \begin{tabular}{lcccc}
    \toprule
Condition & Start Time & End Time & Body Part \\ \hline
sit down on a chair & 0.0 & 5.0 & legs, spine \\
stand up from chair  & 5.0 & 7.0 & legs, spine \\
turn around & 7.0 & 11.5 & legs, spine \\
walk forward & 11.5 & 17.5 & legs, spine \\
5\_stewart\_0\_1\_1,\$785440\$911328 & 5.0 & 12.85 & head, arms \\
5\_stewart\_0\_1\_1,\$918560\$1050080 & 17.5 & 24.3 & head, arms  \\ \bottomrule
    \end{tabular}
    \end{adjustbox}
\end{table}

\noindent \textbf{Explanation:}
\begin{itemize}
    \item Text Description: ``sit down on a chair''
    \begin{itemize}
        \item Start and End Time in Overall Motion: \# 0.0 \# 5.0
        \item Controlled Body Parts: \# legs \# spine
    \end{itemize}
    \item Audio Input File: speak:5\_stewart\_0\_1\_1
    \begin{itemize}
        \item Start and End Frames in Audio File: \$785440\$911328
        \item Start and End Time in Overall Motion: \# 5.0 \# 12.85
        \item Controlled Body Parts: \# head \# arms
    \end{itemize}
\end{itemize}

Notable, the LLM may occasionally produce calculation errors, such as miscalculating the duration of the second audio segment. However, these issues can be easily resolved through code refinement.
The motion generated by our \modelname based on the refined timeline is provided in \texttt{motion-planning-result.mp4}.

\section{Conflict Resolution and Flexibility in Multi-Modal Control}
\label{sup:flexibility}
\modelname’s default configuration assigns upper-body control to audio signals and lower-body control to text descriptions. However, there are scenarios where users may wish to explicitly control specific upper-body motions, such as waving hands while talking. To clarify how our method handles this case, we provide the following example:

\begin{table}[h]
\centering
\scriptsize
\begin{adjustbox}{center}  
\begin{tabular}{lccc}
\toprule
Condition & Start Time & End Time & Body Part \\  
\midrule
a man walks then waves hands & 0.0 & 8.0 & legs, spine \\  
speak:1\_wayne\_0\_103\_103,\$9248\$99808 & 2.578 & 8.238 & left arm, right arm, head \\  
speak:1\_wayne\_0\_103\_103,\$107552\$196064 & 9.722 & 15.254 & left arm, right arm, head \\  
\bottomrule
\end{tabular}
\end{adjustbox}
\end{table}

In this example, a conflict arises between the upper-body motion cues from the text description ``waves hands'' and the speech audio during the overlapping period from 2.578 seconds to 8.0 seconds. To resolve such conflicts, our method relies on a predefined timeline that gives precedence to the speech audio in governing upper-body motion during this interval. Consequently, the `waves hands’’ motion is suppressed between 2.578 and 8.0 seconds.

While this illustrates a limitation of our default setup, the flexibility of MOCO offers a way to address such conflicts by enabling finer-grained user control. If users wish to control upper-body motions using text during speech, we can simply adjust the timeline as follows:
\begin{table}[h]
\centering
\scriptsize
\begin{adjustbox}{center}  
\begin{tabular}{lccc}
\toprule
Condition & Start Time & End Time & Body Part \\  
\midrule
a man walks & 0.0 & 8.0 & legs, spine \\  
wave hands & 7.0 & 11.0 & left arm, right arm \\  
speak:1\_wayne\_0\_103\_103,\$9248\$99808 & 2.578 & 8.238 & left arm, right arm, head \\  
speak:1\_wayne\_0\_103\_103,\$107552\$196064 & 9.722 & 15.254 & left arm, right arm, head \\  
\bottomrule
\end{tabular}
\end{adjustbox}
\end{table}

After modifying the timeline, the text command ``wave hands'' controls the arms from 7.0 to 11.0 seconds. Since the speech audio spans from 2.578 to 15.254 seconds, there is an overlap between 7.0 and 11.0 seconds during which both audio and text attempt to control the arms. According to our conflict resolution rules, \textbf{the condition with fewer controlled body parts takes precedence}. In this case, the text command ``wave hands’’ governs the arms during the overlapping period.

This example demonstrates the flexibility of our method. Although our default configuration prioritizes audio for the upper body and text for the lower body, users can easily reconfigure these priorities through timeline adjustments to achieve their desired behavior. The corresponding visualization is provided in the supplementary materials as \texttt{flexibility.mp4}.

\section{Limitations of Weighted Sum in Multi-Modal Motion Generation}
\label{sup:wa}

\begin{figure*}[htbp]
    \centering
    \includegraphics[width=\linewidth, trim={0 0 0 0}, clip]{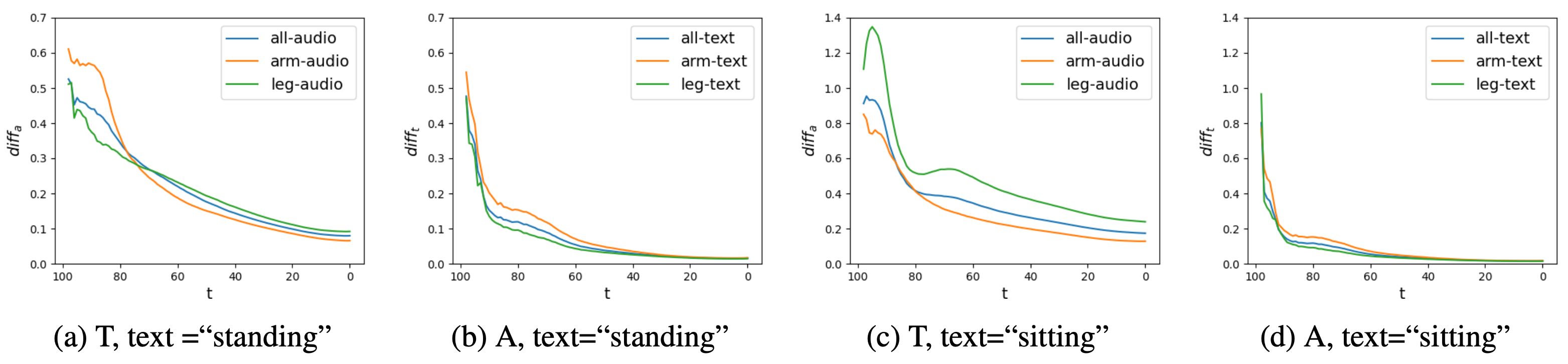}
    \caption{Comparison of differences calculated by the speech-to-gesture and text-to-motion models during motion updates. ``\textbf{T}'' denotes using a text-to-motion model to update motion, while ``\textbf{A}'' denotes using a speech-to-gesture model to update motion. The results show that the speech-to-gesture model computes larger differences than the text-to-motion model, indicating it adjusts the motion more aggressively based on the conditions. This explains why the weighted sum method attains strong performance on speech-to-gesture metrics but poor performance on text-to-motion metrics. Additionally, when the text condition is ``sitting,'' the speech-to-gesture model calculates larger differences in the legs than in the arms. This is counterintuitive and could be attributed to data bias in the speech-to-motion dataset. Best viewed in color.}
    \label{fig:baseline}
\end{figure*}

To understand why the \textit{Weighted Sum} method achieves good results in speech-to-gesture metrics but poor performance in text-to-motion metrics, we conducted the following experiments.

Given both speech and text inputs, we used only the text-to-motion model to update the motion. At each denoising step~$t$, we computed the difference $\text{diff}_t$ between the speech-to-gesture model's prediction—based on the speech input and the current motion from the text-to-motion model—and the current motion from the text-to-motion model. This difference quantifies how much the speech-to-gesture model perceives a mismatch between the speech condition and the current motion. Conversely, when we used only the speech-to-gesture model to update the motion, the calculated difference indicated how much the text-to-motion model perceived a mismatch between the text condition and the current motion. A larger difference suggests a greater mismatch and that the model will update the motion more aggressively.

We recorded these differences in both scenarios and divided them into whole body, arms, and legs for clearer illustration. Comparing \cref{fig:baseline} (a) and (b), as well as \cref{fig:baseline} (c) and (d), we found that the differences calculated by the speech-to-gesture model are larger than those by the text-to-motion model. This indicates that the speech-to-gesture model adjusts the motion more aggressively based on its conditions than the text-to-motion model does. This explains why, when using the weighted sum method, the generated result closely resembles that produced entirely by the speech-to-gesture model.

Furthermore, by comparing \cref{fig:baseline} (a) and (c), which have different text conditions, we observe that when the text condition is ``sitting,'' the differences calculated by the speech-to-gesture model in the legs are larger than in the arms. This is counterintuitive since speech is typically associated with upper-body gestures rather than lower-body movements. Conversely, when the text condition is ``standing,'' the differences in the legs are smaller than in the arms, aligning with expectations. This phenomenon may be attributed to data bias in the speech-to-motion dataset, where most motions are performed in standing positions.

These observations reveal the limitations of weighted sum in multi-modal motion generation and suggest the validity of our proposed decoupled denoising process.

\section{Computational Complexity}
\label{sup:cc}

\begin{table}[ht]
\centering
    \caption{Complexity of each denoiser of \modelname.}
    \label{tab:complexity}
    \begin{tabular}{l|cccc}
    \toprule
         & \makecell{Parameters \\ (M)} & \makecell{Model Size \\(MB)} & \makecell{FLOPs \\(G)} & \makecell{Inference Time \\(ms/frame)} \\ \midrule
    $G_{\text{T2M}}$ & 27.01 & 103.02 & 5.19 & 2.26 \\
    $G_{\text{S2G}}$ & 36.86 & 140.62 & 6.72 & 4.30 \\
    $G_{\text{T2V}}$ & 0.34 & 1.31 & 0.06 & 6.20 \\
    $G_{\text{S2D}}$ & 36.94 & 140.94 & 6.74 & 4.37 \\
    \bottomrule
    \end{tabular}

\end{table}

Our framework, \modelname, comprises four transformer-based denoisers: $G_{\text{T2M}}$ for text-to-motion (T2M), $G_{\text{S2G}}$ for speech-to-gesture (S2G), $G_{\text{T2V}}$ for trajectory-to-velocity (T2V), and $G_{\text{S2D}}$ for speech-to-details (S2D), which manages facial expressions and hand poses. To clearly illustrate the computational complexity of \modelname, we present various metrics, including the number of parameters, model size, FLOPs, and inference time on a single NVIDIA 4090 GPU, as shown in \cref{tab:complexity}.

As indicated in the table, our framework is overall lightweight and sufficiently fast. Specifically, the speech-to-gesture denoiser $G_{\text{S2G}}$ and the speech-to-details denoiser $G_{\text{S2D}}$ are relatively larger than the other denoisers due to additional cross-attention parameters. In contrast, the trajectory-to-velocity denoiser $G_{\text{T2V}}$ is the most lightweight module, featuring fewer hidden state dimensions and transformer layers because the task it handles involves low-dimensional data. However, the introduction of a guidance mechanism for more accurate predictions results in $G_{\text{T2V}}$ having the longest inference time.

Finally, to generate the motion sequences for a 35-second demo video consisting of nine clips under different conditions and with a total duration of 54 seconds, our method completed the body motion generation task in only 3.72 seconds. This fast generation time highlights the potential of our approach for real-time applications.

\rev{For a like-for-like comparison against competing methods, we additionally report end-to-end generation speed (seconds per sequence) under the concurrent multi-modal setting in \cref{tab:stmc} (\aref{sup:stmc}): \modelname is the fastest at $0.69$\,s/seq, ahead of STMC ($0.75$\,s/seq) and the StableMoFusion-backbone variant ($0.86$\,s/seq), confirming that \modelname attains its naturalness and synchronization advantages without incurring additional inference cost.}

\section{Additional Experiments}

\begin{table}[ht]
\centering
\caption{\rev{Comparison with per-step composition (STMC) and a stronger diffusion backbone (StableMoFusion). Best among generative methods is in \textbf{bold}.}}
\label{tab:stmc}
\resizebox{\linewidth}{!}{%
\setlength{\tabcolsep}{6pt}
\begin{tabular}{l|cc|cc|cc|c}
\toprule
 & \multicolumn{2}{c|}{Naturalness} & \multicolumn{2}{c|}{Text2Motion} & \multicolumn{2}{c|}{Speech2Gesture} & Speed \\
Method & MC$\uparrow$ & Jitter$_{\text{spine}}\downarrow$ & FID+$\downarrow$ & R1$\uparrow$ & FID-A$\downarrow$ & BC$\uparrow$ & s/seq$\downarrow$ \\
\midrule
GT             & $-1.95$ & 0.59 & 0.000 & 40.0 & --   & --   & --   \\
\modelname{} (Ours) & $\mathbf{-3.39}$ & \textbf{0.60} & 0.862 & 24.6 & \textbf{3.83} & 2.72 & \textbf{0.69} \\
STMC~\cite{petrovich2024multi}          & $-3.84$ & 0.63 & \textbf{0.684} & \textbf{27.5} & 4.44 & 2.22 & 0.75 \\
StableMoFusion~\cite{huang2024stablemofusion} & $-3.91$ & 0.69 & 0.752 & 27.2 & 4.83 & \textbf{2.92} & 0.86 \\
\bottomrule
\end{tabular}%
}
\end{table}

\subsection{\rev{Comparison with STMC and Stronger Backbones}} \label{sup:stmc}
\rev{This section presents experiments comparing \modelname against two additional baselines, with results reported in \cref{tab:stmc}. We include STMC~\cite{petrovich2024multi} because its single-modality-per-step composition inspired our multi-modal decoupled denoising, and StableMoFusion~\cite{huang2024stablemofusion} to examine whether a backbone that is stronger on text-to-motion transfers into improved motion quality in our concurrent control setting. To adapt STMC to the multi-modal setting while faithfully preserving its original design, we use self-attention to process both text and audio conditions, share parameters across denoisers, and rely on an LLM to automatically annotate the assignment of body parts. In addition to the text-to-motion (FID+, R1) and speech-to-gesture (FID-A, BC) metrics used in the main paper, we further report two naturalness-oriented metrics: \emph{MotionCritic} (MC)~\cite{wang2025aligning}, a learned motion-quality score aligned with human perception, and $\text{Jitter}_{\text{spine}}$, the jerk of the spine joints at the boundary between the audio-controlled upper body and the text-controlled lower body, which quantifies how smoothly the two independently denoised regions are stitched together.}

\begin{table}[ht]
\centering
\caption{\rev{Evaluation of trajectory control under classifier-free guidance (CFG), OmniControl (OC)~\cite{xie2023omnicontrol}, and post-hoc L-BFGS optimization, on location (m) and orientation (rad). OmniControl is a dedicated spatial-control method used to replace our trajectory-to-velocity module $G_{\text{T2V}}$; the first three rows therefore use $G_{\text{T2V}}$.}}
\label{tab:trajecotry}
\begin{tabular}{ccc|cc|cc}
\toprule
\multicolumn{3}{c|}{Method} & \multicolumn{2}{c|}{Location} & \multicolumn{2}{c}{Orientation} \\
CFG & OC & L-BFGS & \makecell{Average\\ Difference} &  \makecell{Goal \\Difference} & \makecell{Average \\Difference} &  \makecell{Goal \\Difference} \\ \hline

\ding{55} & \ding{55} & \ding{55}   & 0.56 & 1.21 & 0.71 & 1.26 \\
\checkmark & \ding{55} & \ding{55}  & 0.57 & 1.32 & 0.83 & 1.51 \\
\checkmark & \ding{55} & \checkmark  & 0.11 & 0.20 & 0.71 & 1.28 \\
\ding{55} & \checkmark & \ding{55}   & 0.44 & 0.79 & 0.53 & 0.97 \\
\ding{55} & \ding{55} & \checkmark   & \textbf{0.08} & \textbf{0.13} & 0.59 & 1.10 \\
\ding{55} & \checkmark & \checkmark  & 0.14 & 0.20 & \textbf{0.48} & \textbf{0.87} \\

\bottomrule
\end{tabular}
\end{table}

\rev{We report the results transparently. On pure text-to-motion fidelity, STMC is in fact slightly stronger than \modelname, and StableMoFusion is likewise competitive on these metrics. However, \modelname achieves better speech-to-gesture fidelity, boundary smoothness, and MotionCritic score. These results suggest that, rather than being uniformly best on every isolated single-modality metric, \modelname delivers the most natural and best-synchronized motion under joint text-and-speech control, as further corroborated by the pairwise user study in \aref{sup:user_addi}. We note that MC should be interpreted only as an auxiliary indicator, because it is trained predominantly on text-driven motion, whereas \modelname produces dense, expressive co-speech upper-body gestures that may lie partly outside its training distribution.}

\subsection{Evaluation of Trajectory Control}

\cref{tab:trajecotry} evaluates trajectory control methodologies by assessing the effects of classifier-free guidance (CFG), OmniControl (OC)~\cite{xie2023omnicontrol}, and L-BFGS optimization on both location (meters) and orientation (radians). Here, OmniControl is a dedicated spatial-control method that replaces our lightweight trajectory-to-velocity module $G_{\text{T2V}}$, whereas the other configurations build on $G_{\text{T2V}}$. For each category, two metrics are reported: Average Difference, the mean deviation between the generated trajectory and the ground truth (GT), and Goal Difference, the discrepancy at the final point relative to the GT.

\rev{The post-hoc L-BFGS optimization is the single most effective component: applied on top of $G_{\text{T2V}}$, it drastically reduces the location error while also modestly improving orientation. In contrast, CFG does not help and can even interfere with the optimization, slightly worsening both location and orientation. Replacing $G_{\text{T2V}}$ with OmniControl improves orientation, and OC$+$L-BFGS achieves the overall best orientation---but at the cost of a larger location error and additional inference time. We therefore retain $G_{\text{T2V}}+$L-BFGS as our default for its better location accuracy and efficiency, and note that adopting a dedicated spatial controller is a worthwhile direction when orientation fidelity is prioritized.}

\rev{We further emphasize, in the interest of transparency, that the absolute orientation accuracy of \modelname{} is still not fully satisfactory. We attribute this to a deliberate design trade-off: \modelname{} adopts a motion representation chosen to facilitate temporal stitching, which is prone to accumulated error along the trajectory---hence the orientation error remains comparatively large even after optimization.}

\subsection{Single Modality Performance}
\begin{table}[ht]
\centering
    \caption{Quantitative results of text-to-motion generation on the HumanML3D test set.}
    \label{tab:ml3d}
    \begin{tabular}{l|ccc|cccc}
\toprule
\multirow{2}{*}{Method} & \multicolumn{3}{c|}{R-Precision} & \multirow{2}{*}{FID$\downarrow$} & \multirow{2}{*}{MM Dist$\downarrow$} & \multirow{2}{*}{Diversity$\uparrow$} & \multirow{2}{*}{MM$\uparrow$} \\

& {Top 1} & {Top 2} & {Top 3}  \\
\midrule
Ground Truth & 0.511 & 0.703 & 0.797 & 0.002 & 2.974 & 9.503 & - \\
T2M-GPT \cite{zhang2023t2mgpt} & 0.491 & 0.680 & 0.775 & 0.116 & 3.118 & 9.761 & 1.856 \\
MDM \cite{tevet2022mdm} & - & - & 0.611 & 0.544 & 5.566 & 9.559 & 2.799 \\
FineMoGen \cite{zhang2023finemogen} & 0.504 & 0.690 & 0.784 & 0.151 & 2.998 & 9.263 & 2.696 \\
MoMask \cite{guo2024momask} & 0.521 & 0.713 & 0.807 & 0.045 & 2.958 & - & 1.241 \\
LMM-Tiny \cite{zhang2024large} & 0.496 & 0.685 & 0.785 & 0.415 & 3.087 & 9.176 & 1.465 \\
LMM-Large \cite{zhang2024large} & 0.525 & 0.719 & 0.811 & 0.040 & 2.943 & 9.814 & 2.683 \\
SynTalker \cite{chen2024enabling} & 0.375 & 0.564 & 0.681 & 4.385 & 4.499 & 9.374 \\
\textbf{\modelname} (Ours) & 0.434 & 0.618 & 0.720 & 0.530 & 3.563 & 9.856 & 2.663 \\
\bottomrule
    \end{tabular}

\end{table}
    \begin{table}[ht]
\centering
    \caption{Quantitative results of speech-to-gesture generation on the BEATX test set.}
    \label{tab:beatv2}
    \begin{tabular}{l|ccccc}
    \toprule
        Method & FGD↓ & BC & Diversity↑ & MSE↓ & LVD↓ \\ \midrule
        FaceFormer \cite{fan2022faceformer} & - & - & - & 7.787 & 7.593 \\
        CodeTalker \cite{xing2023codetalker} & - & - & - & 8.026 & 7.766 \\ 
        S2G \cite{ginosar2019learning} & 28.15 & 4.683 & 5.971 & - & - \\
        Trimodal \cite{yoon2020trimodal} & 12.41 & 5.933 & 7.724 & - & - \\ 
        HA2G \cite{liu2022learning} & 12.32 & 6.779 & 8.626 & - & - \\ 
        DisCo \cite{liu2022disco} & 9.417 & 6.439 & 9.912 & - & - \\ 
        CaMN \cite{liu2022beat} & 6.644 & 6.769 & 10.86 & - & - \\ 
        DiffStyleGesture \cite{yang2023diffusestylegesture} & 8.811 & 7.241 & 11.49 & - & - \\
        TalkShow \cite{yi2023talkshow} & 6.209 & 6.947 & 13.47 & 7.791 & 7.771 \\ 
        EMAGE \cite{liu2023emage}    & 5.512 & 7.724 & 13.06 & 7.680 & 7.556 \\ 
        ProbTalk \cite{liu2024probtalk} & 6.170 & 8.099 & 10.43 & 8.990 & 8.385 \\ 
        SynTalker \cite{chen2024enabling} & 6.413 & 7.971 & 12.72 & - & - \\
        \textbf{\modelname} (Ours) & 5.543 & 7.089 & 14.05 & 7.285 & 7.573 \\ \bottomrule
    \end{tabular}

\end{table}

To demonstrate \modelname's performance in single-modality scenarios, we trained it from scratch on HumanML3D for text-to-motion and on BEAT2 for speech-to-gesture, respectively, ensuring a fair comparison. The results, presented in \cref{tab:ml3d} and \cref{tab:beatv2}, show that in the HumanML3D text-to-motion benchmark (\cref{tab:ml3d}), our model achieves performance comparable to the widely-used MDM. This outcome is expected since our text-to-motion denoiser, $G_{\text{T2M}}$, is based on MDM. In the BEAT2 speech-to-gesture benchmark (\cref{tab:beatv2}), \modelname attains competitive performance compared to state-of-the-art methods.

\subsection{Details of User Study}
\label{sup:user}

To construct the questionnaire for the user study, we first generated 20 videos using each method, including \modelname (Ours), Pseudo-Text, Weighted Sum, and Combine Once. The videos generated by \modelname were then concatenated with the other videos, resulting in 60 pairs of comparison videos. The user study involved 10 participants, each of whom evaluated all 60 pairs of comparison videos. Participants were asked to select their preferred video or indicate that neither was better based on one of the following criteria: text following, beat synchronization, body coherence, and temporal fluidity. The first 40 video comparisons involved \modelname versus Pseudo-Text and \modelname versus Weighted Sum, focusing on text following and beat synchronization. The last 20 video comparisons involved \modelname versus Combine Once, focusing on body coherence and temporal fluidity. The statistical results are presented in the main paper.

\subsection{\rev{Extended Pairwise-Preference Study against STMC.}}
\label{sup:user_addi}
\rev{The study above compares \modelname against the weighted-sum, pseudo-text, and combine-once baselines. To further corroborate the metric-level comparison with STMC reported in \aref{sup:stmc} through human perception, we additionally conducted a larger-scale pairwise-preference study against STMC~\cite{petrovich2024multi} (the same adapted STMC baseline as in \aref{sup:stmc}). This study involved \textbf{51 participants} drawn from a bachelor's-level (undergraduate) background, none of whom were involved in the project; each evaluated 60 comparison videos. We report participant background explicitly here for transparency, as the demographic composition of evaluators can influence perceptual judgments. For each video pair, participants chose the preferred clip---or ``no preference''---along three criteria: text alignment, audio synchronization, and naturalness.}

\rev{The results are summarized in \cref{tab:user_ext}. \modelname is preferred over STMC on all three criteria, with the margin reaching statistical significance for \emph{audio synchronization} ($47.1\%$ vs.\ $39.4\%$, $p=0.02$) and \emph{naturalness} ($46.1\%$ vs.\ $39.8\%$, $p=0.03$), using a two-sided binomial test with ties excluded. For \emph{text alignment} the preference favors \modelname but does not reach significance ($45.9\%$ vs.\ $42.0\%$, $p=0.11$), which is consistent with the metric-level observation in \aref{sup:stmc} that STMC is competitive on text-to-motion metrics. Taken together, these perceptual results reinforce our central claim: under concurrent text-and-speech control, \modelname produces motion that human viewers judge to be better synchronized and more natural, even where it does not dominate on isolated text-to-motion metrics.}

\begin{table}[ht]
\centering
\caption{\rev{
    Extended pairwise-preference user study comparing \modelname{} against STMC~\cite{petrovich2024multi} under concurrent text-and-speech control. Each cell reports the fraction of comparisons in which participants preferred \modelname{}, expressed no preference, or preferred STMC. The $p$-value is from a two-sided binomial test on the preference between the two methods (ties excluded); $^{*}$ denotes statistical significance at $p<0.05$.
}}
\label{tab:user_ext}
\begin{tabular*}{\linewidth}{@{\extracolsep{\fill}}lcccc}
\toprule
Criterion & Prefer \modelname{} & No Preference & Prefer STMC & $p$-value \\
\midrule
Text Alignment      & \textbf{45.9\%} & 12.1\% & 42.0\% & 0.11 \\
Audio Synchronization & \textbf{47.1\%} & 13.5\% & 39.4\% & 0.02$^{*}$ \\
Naturalness         & \textbf{46.1\%} & 14.1\% & 39.8\% & 0.03$^{*}$ \\
\bottomrule
\end{tabular*}
\end{table}

\section{Details of Multi-Modal Benchmark}
\label{sup:benchmark}

To effectively evaluate our proposed task, we developed a multi-modal benchmark comprising 1,000 test clips, following the methodology outlined in \cite{petrovich2024multi}. Each test clip is automatically generated and includes two text descriptions and two audio clips.

For the text descriptions, we manually curated a set of 40 texts focusing on lower-body movements commonly associated with speech delivery or conversation. These descriptions provide the necessary context for evaluating the corresponding movements within the benchmark.
Regarding the audio clips, we selected recordings from the BEAT2 dataset, specifically choosing eight speakers with speaker IDs below 10. These audio files were segmented into clips using a Voice Activity Detector (VAD), resulting in 694 audio clips with an average duration of 9.14 seconds.

The 1,000 test clips were generated through an automated process that utilizes the curated text descriptions and audio clips. For each test clip, two text descriptions are randomly selected and assigned random durations. Subsequently, two neighboring audio clips are randomly chosen. The start times for both the text and audio intervals are determined randomly, allowing the sequence to commence with either text or audio. This process results in the creation of four intervals that correspond to the selected text descriptions and audio clips.

Optional text descriptions are listed here:

\begin{lstlisting}
  walk in a circle clockwise 
  walk in a circle counterclockwise 
  walk in a quarter circle to the left 
  walk in a quarter circle to the right 
  turn 180 degrees to the left on the left foot 
  turn 180 degrees to the left on the right foot 
  turn left 
  turn right 
  walk forwards 
  walk backwards
  slowly walk forwards 
  slowly walk backwards 
  quickly walk forwards 
  quickly walk backwards 
  run 
  jogs forwards  
  jogs backwards  
  slowly walk in a circle  
  perform a squat  
  sit down  
  turn around then sit down in a chair  
  sit down then get back up and walk back  
  sit down for a moment  
  step back and sit down  
  sit down indian style  
  take a step to their right and sit down  
  sit criss cross  
  sit down on the ground and cross their legs  
  squat down  
  sit on a high object  
  sit on a barstool and rest their legs on the stool  
  take a large step and sits on a stool  
  get down on their knees  
  sit on the ground with his legs extended in front of him  
  walk up to a backwards chair and sit down on it with legs outstretched  
  sit down and adjust themselves  
  sit down and swap their legs crossing back and forth  
  sit and lie down on a lounge chair  
  sit down and lean on the chair  
  sits very still in the chair
\end{lstlisting}

\end{document}